\documentclass{SPAICE}
\usepackage{subcaption}
\usepackage{multirow}
\usepackage{booktabs}
\usepackage{comment}

\def\authorEmail{jd.norberto@campus.fct.unl.pt}
\def\AuthorShort{J. Norberto et al.}

\author[1]{Jo\~{a}o Norberto\thanks{Corresponding author. E-Mail: \authorEmail}}
\author[1]{Ricardo N. Ferreira}
\author[1]{Cl\'{a}udia Soares}
\affil[1]{NOVA School of Science and Technology, Caparica, Portugal}

\title{Online Learning for Dynamic Constellation Topologies}

\begin{document}

\maketitle

\begin{abstract}
The use of satellite networks has increased significantly in recent years due to their advantages over purely terrestrial systems, such as higher availability and coverage. However, to effectively provide these services, satellite networks must cope with the continuous orbital movement and maneuvering of their nodes and the impact on the network's topology. 
In this work, we address the problem of (dynamic) network topology configuration under the online learning framework. As a byproduct, our approach does not assume structure about the network, such as known orbital planes (that could be violated by maneuvering satellites). We empirically demonstrate that our problem formulation matches the performance of state-of-the-art offline methods. Importantly, we 
demonstrate that our approach is amenable to constrained online learning, exhibiting a trade-off between computational complexity per iteration and convergence to a final strategy.
\end{abstract}

\section{Introduction}

From individual satellites to sophisticated constellations offering global (or nearly global) coverage, satellite technology has advanced significantly from the early days of space exploration. Depending on the orbital altitude in which they are deployed, satellites have different use cases. E.g., geostationary satellites are used for weather forecasting~\cite{hertzfeld2004weather}, as they are capable of providing real-time data. Contrastingly, Low Earth Orbit (LEO) satellites have higher orbital speed but allow for real-time communication due to low propagation delays, high bandwidth, and higher throughput when compared with other orbital altitudes~\cite{papapetrou2007distributed, han2021dynamic}, thus being used, e.g., for internet provision~\cite{shaengchart2023starlink, de2015satellite, chen2024concept}. The characteristics of LEO satellites, as well as the increasing importance of satellite communications to modern society, justify the growing interest in LEO constellations.

The maintenance of satellites and their respective constellations is crucial to preserve the services that are widely used by Earth's population every single day. The growth of these constellations~\cite{9461407, kulu2024satellite}, with the limited lifespan of the satellites~\cite{gonzalo2014challenge}, poses a challenge regarding the need to reconfigure their network every time a new satellite is added or removed. Furthermore, in the last decade, the problem of satellite collisions has become increasingly relevant, as more satellites are being launched into space, creating more debris and increasing the probability of collisions, as pointed out by Donald Kessler~\cite{kessler1978collision}. These factors can potentially impact the consistency and quality of the services provided~\cite{guo2025resilience}, leading to both social and economic consequences. It is important that the topology of their networks dynamically adapt to these changes, allowing for stable and more robust services. Nevertheless, a purely dynamic strategy, where satellites would acquire information about the environment and their neighbors every time a new packet needs to be sent, would not be desirable due to the impact it would have on performance. Hence, there needs to be a trade-off between the performance of these methods and adaptation to the dynamic behavior of the network's topology.

\paragraph*{Contributions.} \label{sec:contributions}
Given the dynamic and critical nature of constellation topologies, we address the problem within the constrained online learning framework. As a first step, we formulate the problem as a novel convex optimization problem. As a byproduct, we do not assume structure for the network, such as known orbital planes, that could be violated by maneuvering satellites.  We demonstrate that our approach is amenable to constrained online learning, exhibiting a trade-off between computational complexity per iteration and convergence to a final strategy. 


\section{Related Work \label{sec:sota}}

The topology of a network of satellites can be seen as the connection map between satellites. To establish the topology of an arbitrary network of satellites, two approaches can be used~\cite{xiaogang2016survey}: static approaches, which discretize the dynamic behavior of a satellite network, and dynamic approaches, which account for changes in the topology and directions of inter-satellite links (ISLs).

Static approaches can be divided into virtual topology (VT) \cite{sun2020improved, lu2013virtual, 10436098} and virtual node (VN) \cite{ekici2002distributed, 10404740}. In VT, the system period of the satellite network is divided into $n$ time slices, where the topology of the network is considered fixed, possibly generating a substantial amount of time slices, which is not desired, as satellites have limited storage. VN assumes that Earth is covered by a network of logical satellite locations, each attributed to the closest real satellite, fixing the topology of the network. Afterwards, to establish the ISLs, multiple strategies can be employed. +Grid is the most commonly used strategy, by connecting each satellite with its four closest neighbors, where two are from the same orbital plane and two from adjacent orbital planes. Other works have tried to improve on +Grid, such as $\times$Grid \cite{mclaughlin2023grid}, which proposes a location-oriented design, simplifying client handovers, or the method proposed by Bhattachejee and Singla \cite{Network-topology-design-at-27k-km/hour}, which exploits repetitive patterns in satellite networks to reduce traffic and latency. Despite simplifying the topology of satellite networks, these approaches do not take into account their dynamics, hence leading to problems due to unexpected changes in the network, e.g., link failures and congestion or topology changes.

Alternatively, dynamic approaches \cite{papapetrou2007distributed, 9931973, leyva2021inter, ron2025time, 8688478} consider the dynamics of satellite networks by allowing satellites to acquire information regarding the topology of the network and the status of the ISLs, leading to improvement in the reliability of routing paths. Han et al.\cite{9931973} leverage the predictability and periodicity of satellite networks and evolution graph theory to create an evolution graph that is weighed by the characteristics of the ISLs. Leyva-Mayorga et al.\cite{leyva2021inter} proposed a framework that addresses the topology problem by solving the inter-plane satellite matching problem, a many-to-many maximum weighted matching problem, where they find the set of feasible pairs of satellites that maximizes the sum of signal-to-noise ratio rates. Papapetrou et al.\cite{papapetrou2007distributed} developed an on-demand protocol that consists of three processes: request area formation, path discovery, and route entry management. These three processes are responsible for defining the minimum route request area, message flooding on the valid routes, and route entry management, respectively. Ron et al.\cite{ron2025time} model the topology of the network as a dynamic time-expanded graph and formulate the problem of ISL establishment as an optimization problem that aims to maximize network capacity while minimizing both link churn and latency. Li et al.\cite{8688478} propose an algorithm that periodically collects information of neighbor satellites, e.g., queuing delay and transmission rate. Initially, each satellite broadcasts a packet containing information about its own characteristics. The neighbors collect that information and mark the sender as an alive neighbor. The time to live of these packets is defined to be three broadcast periods.

\section{Online Learning \label{sec:oco_introduction}}

Online Learning, also referred to as Online Convex Optimization (OCO), can be seen as a structured repeated game, where a player iteratively makes decisions~\cite{hazan2016introduction,orabona2019modern}. 
A player must iteratively choose an action over some convex set $\mathcal{K}$, while an adversary selects a convex cost function $f_t$ \emph{a posteriori}, causing the player to incur a cost associated with their decision. The objective of the player is to perform as well as a set of comparator decisions for all the rounds of interaction. The metric the player intends to minimize is the regret, which is defined as the difference between the cumulative cost incurred by the player and the cumulative cost incurred by the comparator, i.e.,
\begin{equation}
\label{eq:regret-definition}
\begin{aligned}
    \text{Regret}_T = \sum_{t=1}^{T} f_t(x_t) - \sum_{t=1}^{T} f_t(u_t) ,
\end{aligned}
\end{equation}
where $x_t$ denotes the actions selected by the player at each time step $t$, and $u_t$ denotes the actions selected by the comparator strategy. When the comparator strategy selects the best fixed action in hindsight, i.e, $u_t = \underset{x \in \mathcal{K}}{\arg\min} \sum_{t=1}^T f_t(x)$, the metric is called the \emph{static regret}. On the other hand, when, at each iteration, the comparator selects the best decision, i.e., $u_t = \underset{x \in \mathcal{K}}{\arg\min} \ f_t(x)$, then the metric is called \emph{dynamic regret}. In this work, we focus on the performance against a dynamic comparator strategy.

To tackle OCO problems, two types of algorithms can be used. Projection-based algorithms are a class of algorithms that project the iterate onto the decision set at each timestep, while projection-free algorithms avoid this possibly expensive computation, since the iterate always lies within the decision set, by using, e.g., linear optimization. Depending on the problem, the projection step can be computationally heavy, and so the benefits of using projection-free algorithms depend on the problem at hand \cite{hazan2016introduction}.

\section{Problem Formulation \label{sec:prob_formulation}}

As a means to setup the Constrained Online Learning problem in the OCO framework detailed above, we formulate the problem of finding a satellite network topology as a weighted least squares, modeling the network topology as a Laplacian matrix (for a detailed explanation of the Laplacian matrix and its properties, see~\cite{bapat2010graphs}). Hence, we formulate the problem of finding a topology $X$ for a set of satellites and base stations, $\mathcal{S}$ and $\mathcal{B}$, respectively, given their positions, viz:\begin{align}
    &\operatorname*{arg\,min}_{X} \; \|(X-P) \odot U\|_F^2 - \operatorname{Tr}(\Lambda \odot X) 
    \label{eq:topology_objfun}\\
    &\text{s.t.} \quad X=X^T,    \label{eq:topology_c1}\\
    &\qquad \sum_{j=1}^n x_{ij} = 0, \quad \forall i \in \{1,2,\cdots, n\}, \label{eq:topology_c3}\\
    &\qquad 0 \leq x_{ii} \le \max\nolimits_{ISL}, \quad \forall i \in \mathcal{S}, \label{eq:topology_c5}\\
    &\qquad 0 \leq x_{ii} \le \max\nolimits_{BSL}, \quad \forall i \in \mathcal{B}, \label{eq:topology_c6}\\
    &\qquad -1 \le x_{ij} \le 0, \quad \forall i \neq j \label{eq:topology_c7}
\end{align} where $\max_{ISL}$ and $\max_{BSL}$ are the maximum number of connections a satellite and base station can have, respectively, $\odot$ represents the element-wise multiplication, $P$ represents the connectivity matrix, $U$ represents the utility, $\|\cdot\|_F$ is the Frobenius norm, and $\Lambda$ is a diagonal matrix used to encourage the connection between satellites and base stations.

The problem of finding a topology given the positions of all satellites is modeled as a weighted least squares where we minimize the Frobenius norm between the Laplacian matrix $X$, representing the topology of the network, and the connectivity matrix $P$. This matrix is composed of either zeros or minus ones. If two satellites cannot connect, their respective entry is zero. Otherwise, it has a value of minus one to encourage connections. Later in this section, we present two approaches to model these connectivity constraints.


By considering a utility matrix $U$, we encourage certain connections by assigning different utilities to each individual ISL. This matrix can encompass metrics such as the packet drop probability \cite{taleb2009} and buffer queue sizes \cite{9931973} to allow for more reliable ISL establishment. In this work, we adopt the reciprocal of the inter-satellite distance as the utility metric, thereby promoting stronger connectivity among neighboring satellites. As in most cases base stations are farther away from satellites than their closest neighbors, we increase their utility by some factor to encourage connections between orbital and terrestrial nodes.



To encourage the connectivity of the network, we add the trace to the objective function \cref{eq:topology_objfun}, as $-\text{Tr}(\Lambda \odot X)$, since the trace of a Laplacian matrix is equal to $2m$, where $m$ is the number of edges of the graph \cite{bapat2010graphs}. The matrix $\Lambda$ is a diagonal matrix that contains two values, $\lambda_{bs}$, the weight given to the base station's connections, and $\lambda_{sat}$, the weight given to the satellite's connections. E.g., in the case of a Laplacian matrix where the first two rows represented base stations and the remaining represented satellites, this matrix would be represented as $\Lambda = \text{diag}( \lambda_{bs}, \lambda_{bs}, \lambda_{sat}, \lambda_{sat})$. Additional properties of a Laplacian matrix are added as constraints. The matrix is symmetric for an undirected graph (\cref{eq:topology_c1}). The sum of its rows is zero (\cref{eq:topology_c3}). The maximum diagonal entry is the maximum number of edges each vertex can have (\cref{eq:topology_c5} and~\cref{eq:topology_c6}), for satellites and base stations, respectively. The off-diagonal values of a Laplacian matrix, for an unweighted graph, are bounded between minus one and zero (\cref{eq:topology_c7}).

\paragraph*{Connectivity Constraints. \label{par:net_top_constr}}

\begin{figure}[h!]
    \centering
    \includegraphics[width=0.55\linewidth]{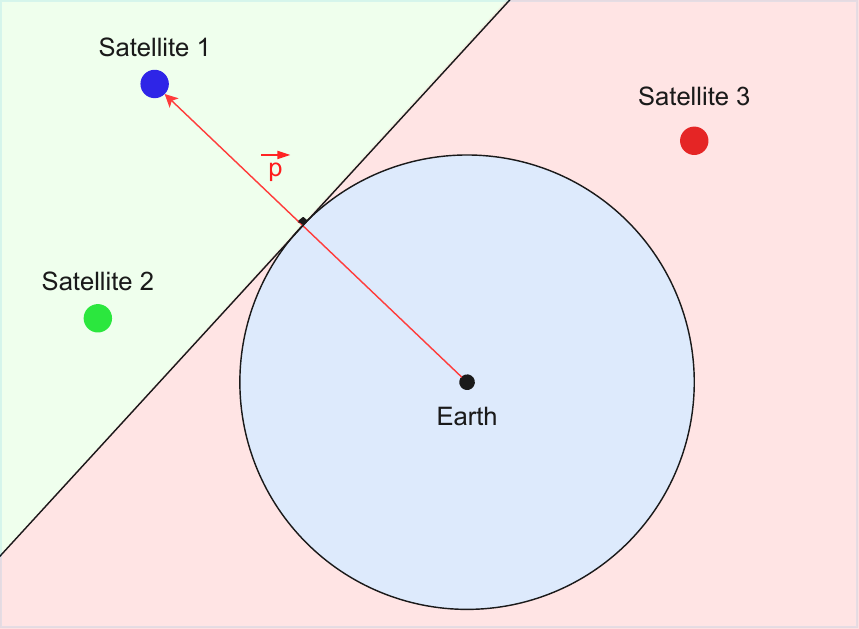}
    \caption{An example of the use of a hyperplane to simulate the field of view (FOV) of a satellite. The area colored in green represents the FOV of satellite 1, while the area colored in red represents the area outside of the satellite's FOV. If a satellite falls within the green area, it can connect with satellite 1.}
    \label{fig:2d_hyperplane}
\end{figure}

With the positions of all satellites in the network, we know that some of these satellites cannot physically connect with others by approximating the field of view of the satellite. Thus, we can enforce some of the entries in the connectivity matrix $P$ to be zero, i.e., all the pairs of satellites $(i, j)$ that cannot have an edge between them. To create the non-connectable set of pairs $(i, j)$, we developed an approximation that uses hyperplanes to define which pair of satellites belongs to each of the non-connectable sets of each satellite. As more hyperplanes are added, the better the field of view of a satellite is approximated, but the slower the method will be. The first approximation we consider to simulate the field of view of a satellite is the use of a single hyperplane. In this case, the hyperplane must be tangent to the surface of the Earth, and the normal vector is defined by the normalized vector of the relative position of the satellite with respect to Earth. Subsequently, the set corresponds to all satellites that are located on the opposite side of the hyperplane, relative to satellite $i$. In~\cref{fig:2d_hyperplane}, an example of this procedure can be seen for satellites in two dimensions.

\begin{figure}[h!]
    \centering
    \includegraphics[width=0.55\linewidth]{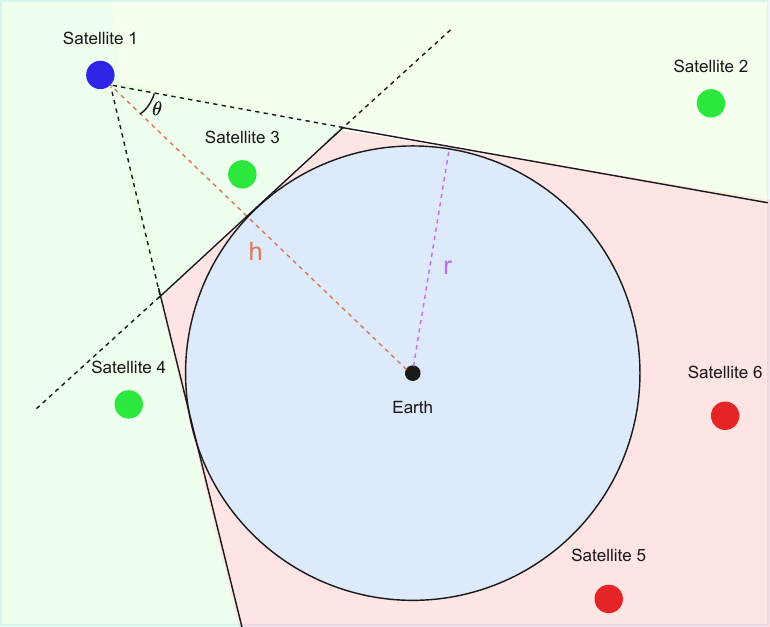}
    \caption{An example of the use of both a hyperplane and a conical surface to approximate the FOV of a satellite. The area colored in green represents the FOV of satellite 1, while the area colored in red represents the area outside of the satellite's FOV. If a satellite falls within the green area, it can connect with satellite 1.}
    \label{fig:combined_preprocessing_approach}
\end{figure}

In addition to the hyperplane, we also consider a conical surface with apex on the satellite in question that is tangent to Earth and contains it. We know that if a line is tangent to a sphere, it is always perpendicular to its radius. Thus, we can find the angle of the solid of revolution generated by that line. We can then compute which satellites belong to the non-connectable set by simply comparing dot products. By combining both methods, we obtain the second approximation to the field of view of a satellite. In this case, we simply do an intersection between the sets obtained from each method. The procedure when considering three dimensions is similar. \cref{fig:combined_preprocessing_approach}, shows an example of this approximation.

\section{Results \label{sec:results}}
To empirically examine the advantages of our method, we first establish the validity of our formulation in~\cref{sec:prob_formulation} and then compare the OCO algorithms' performance.

\paragraph*{Validity of the proposed formulation.}\label{sec:results-sota}

In this section, we compare the offline version of our method using the reciprocal of the distance between satellites as a utility measure with the state-of-the-art approach +Grid. The diagonal entries of the matrix $\Lambda$ have a weight of $10^{2}$ to encourage a connected topology. To compare both approaches, we consider an Iridium-like constellation with six elliptical orbital planes, each with eleven satellites, and no base stations.

\cref{tab:+grid_vs_our_method} summarizes the results of the method +Grid and our approach, considering two different FOV. Note that the comparison is biased towards +Grid, as it has information about orbital planes and so connects a satellite with the two closest neighbors of the same orbital plane and the two closest neighbors from adjacent orbital planes. Nevertheless, despite not using orbital plane information, we see that the results of our approach are comparable to the state-of-the-art, with minor impact on metrics such as average degree, average shortest path, and average clustering coefficient.


\begin{table}[]
\centering
\scriptsize
\caption{Graph statistics for an Iridium-like constellation when using both the +Grid method and our approach. The metrics considered in this table are the total number of edges in the graph, ``E''; the average degree of a satellite, ``A-Deg''; the density of the generated graph, ``D''; the number of connected components, ``CC''; the average shortest path, ``A-SP''; and the average clustering coefficient, ``A-C''.}
\begin{tabular}{@{}ccccccccc@{}}
\cmidrule(lr){2-8}
 & \textbf{FOV:} & \textbf{E $\uparrow$} & \textbf{A-Deg $\uparrow$} & \textbf{D $\downarrow$} & \textbf{CC $\downarrow$} & \textbf{A-SP $\downarrow$} & \textbf{A-C $\downarrow$} \\
\midrule
+Grid & --- & \textbf{130} & \textbf{3.94} & 0.06 & 1 & \textbf{4.04} & \textbf{0.02} \\
\midrule
\multirow{2}{*}{Ours}
 & H  & 121 & 3.67 & 0.06 & 1 & 4.68 & 0.15  \\
 & H+C & 113 & 3.42 & 0.06 & 1 & 4.88 & 0.12  \\
\bottomrule
\end{tabular}
\label{tab:+grid_vs_our_method}
\end{table}

\paragraph*{Comparison of the OCO algorithms' performance. \label{sec:oco_results}}

We tested the performance of two Online Convex Optimization (OCO) algorithms~\cite{hazan2016introduction}: Online Gradient Descent (OGD), a simple projection-based algorithm that adds a projection step onto the decision set after the computations of vanilla gradient descent; and Online Conditional Gradient (OCG), the online version of the Frank-Wolfe algorithm~\cite{hazan2016introduction}. This algorithm eschews the possible expensive projection onto the decision set by iteratively solving linear optimization problems. 

To test both algorithms, we simulate the movement of 18 satellites in a single orbital plane with two fixed base stations for one thousand iterations. The utility measure and the $\Lambda$ matrix used are the same as before. To further enforce the connections between satellites and base stations, each entry of the utility matrix $U$ representing a base station is scaled by $10^{2}$. At iteration zero, only our offline algorithm generates a topology, which is used as input to both online algorithms, conditioning the future topologies that will be generated by both. For consequent iterations, three topologies were generated: one from our offline algorithm, which is generated from scratch at each timestep; one from OGD; and one from OCG. We then compute the average difference per entry between the Laplacian matrices generated by the online algorithms and our offline variant, considering the latter as our gold standard,
\begin{equation}
  \frac{\left\| X_{\text{on}} - X_{\text{off}}\right\|_F^2}
  {\#\left[(x_{ij} \neq 0), \forall_{x_{i, j}} \in X_{\text{off}} \right]},
  \label{eq:plot_formula}
\end{equation}
where $X_{\text{on}}$ represents the Laplacian matrix resulting from the online algorithms, $X_{\text{off}}$ the Laplacian matrix resulting from our offline methodology. The optimization problem in~\cref{sec:prob_formulation} is solved with the help of the Python library CVXPY~\cite{diamond2016cvxpy}.

\begin{figure}[t]
    \centering
    \scriptsize
    \includegraphics[width=.7\columnwidth]{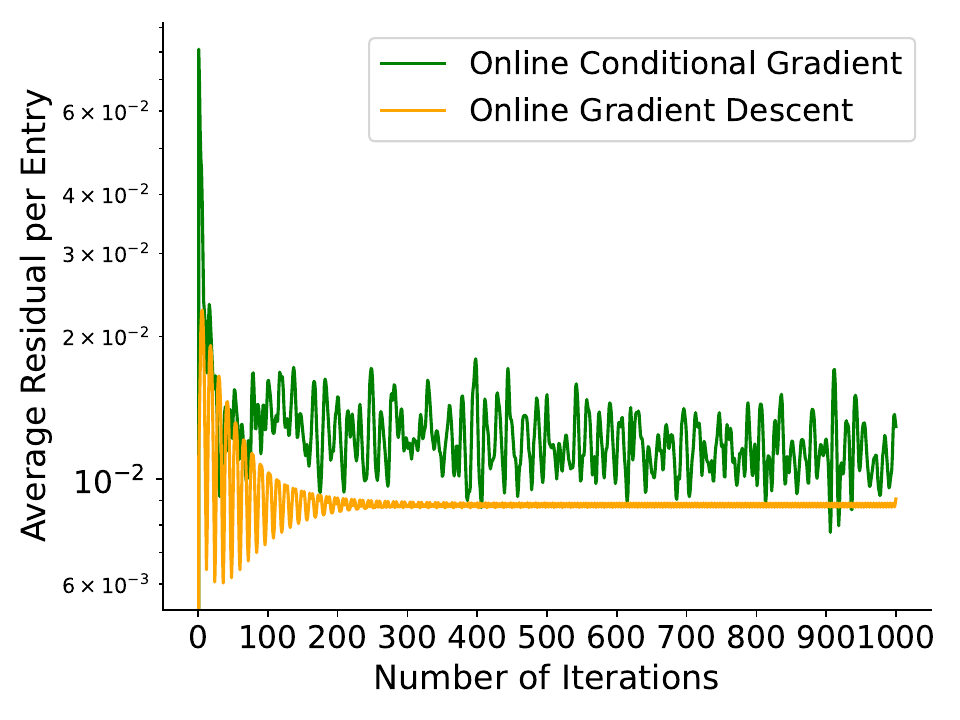}
	\caption{Smoothed average residual per entry between Laplacian matrices produced by the online algorithms and our offline method, following \cref{eq:plot_formula}. The lower the residual, the closer the resulting matrices are to the offline method.}
	\label{fig:oco_results_single_plane}
\end{figure}

\cref{fig:oco_results_single_plane} shows that initially, OCG incurs a higher error, decreasing it over time, with OGD achieving the smallest value. This is expected as OCG trades regret for computational complexity~\cite{hazan2016introduction}. To verify this trade-off, we computed the wall time for OCG and OGD throughout the experiment, yielding the values $2.712_{\pm0.06}$~ms and $3.954_{\pm0.07}$~ms respectively. OCG takes more iterations to learn the network dynamics than OGD, while incurring, on average, 31\% faster per-iteration computations.


\section{Discussion \label{sec:discussion}}

The developed methods show promising results, approximating the results obtained by +Grid without any structure assumption by using our offline formulation and a simple utility measure. Nevertheless, there is space for improvement by considering more meaningful ISL characteristics. Regarding OCO algorithms, online gradient descent showed promising results, as it can model the movement of a single orbital plane by just performing a gradient step and a projection. For future work, we will further analyze and develop OCO algorithms in the context of satellite topology, as well as test different approaches to condition the topologies generated by OCO algorithms. Reinforcement Learning algorithms also surface as a possible solution, allowing the modeling of non-convex optimization functions, despite possibly incurring in catastrophic forgetting.

\clearpage
\section*{Ethics Statement}
This work develops learning-based methods for the adaptive configuration of satellite network topologies to improve the robustness and efficiency of space communication infrastructure. While evaluated solely in simulation and without the use of sensitive data, technologies that optimize satellite networks may have dual-use implications. We therefore emphasize responsible research and deployment aligned with international efforts toward the sustainable, safe, and peaceful use of outer space, in light of the growing congestion in Low Earth Orbit and the need for resilient space infrastructure.


\printbibliography
\addcontentsline{toc}{section}{References}

\end{document}